# Information Fusion for Anomaly Detection with the Dendritic Cell Algorithm

Julie Greensmith, Uwe Aickelin and Gianni Tedesco, School of Computer Science, University of Nottingham

**Abstract**

Dendritic cells are antigen presenting cells that provide a vital link between the innate and adaptive immune system, providing the initial detection of pathogenic invaders. Research into this family of cells has revealed that they perform information fusion which directs immune responses. We have derived a Dendritic Cell Algorithm based on the functionality of these cells, by modelling the biological signals and differentiation pathways to build a control mechanism for an artificial immune system. We present algorithmic details in addition to experimental results, when the algorithm was applied to anomaly detection for the detection of port scans. The results show the Dendritic Cell Algorithm is successful at detecting port scans.

## 1. Introduction

Denritic Cells (DCs) are natural anomaly detectors. In this paper we present a Dendritic Cell Algorithm (DCA) approach to information fusion, combining key elements of immunological theory with the engineering principles of data fusion. In the human immune system, DCs have the power to suppress or activate the immune system by correlation of signals representing their environment, combined with locality markers in the form of antigens. Antigens are proteins in structure and are any protein to which the immune system can potentially respond. These cells are responsible for the detection of pathogens in the human body through the correlation of information (in the form of molecular signals) within the environment. By using an abstraction of DC behaviour, similar detection properties are shown, resulting in an algorithm capable of performing anomaly detection. The resultant algorithm uses a set of weights derived or the processing of input signals from actual immunological data, generated through an interdisciplinary collaboration with immunologists[32].

DCs in particular are suitable as inspiration for intrusion detection for two reasons. Firstly, DCs themselves perform an intrusion detection role within the human immune system. Secondly, DCs perform their function with low rates of false positives and high rates of true positives - properties essential to any anomaly detection technique. In essence, DCs are multisensor data fusion agents through processing environmental molecular signals. This makes them ideal inspiration for the development of a data fusion algorithm.

The DCA was introduced in 2005[9] and has demonstrated potential as a classifier for static machine learning data [9], as a simple port scan detector under experimental conditions[11] and in real time[10]. Our results show that the DCA can successfully detect anomalous processes forming a port scan attack. The DCA is inspired by the human immune system and is termed an artificial immune system (AIS). While the majority of AIS algorithms do not perform data fusion, idiotypic network models are used for the purpose of robotic control [12]. Although belonging to the field of artificial immune systems, the DCA differs from other immune inspired anomaly detection algorithms in a number of significant ways:
- The algorithm is based on cutting edge experimental immunology.
- DCs combine multiple signals to assess the current context of their environment.
- Asynchronolsly DCs sample another data stream (antigen) to be combined with the fused signals.
- The correlation between context and antigen leads to the detection of anomalies.
- Unlike other anomaly detection algorithms, there is no pattern matching based on string similarity metrics.

The aims of this paper are threefold: to model artifical DCs drawing inspiration from the DCs of the human immune system; to present a resultant algorithm through a formalised description; and to apply the algorithm to an example anomaly detection problem. As this algorithm is a novel algorithm, it is not yet fully characterised. As a result, fine grained analysis of the selection of weights and comparison to other standard techniques are not discussed in this paper. Please refer to [8] for further experiments.

In this paper The DCA is applied to the detection of a port scan, which forms a convenient small-scale computer security problem. Section 2 contains relevant background information regarding the problem of port scans and current scanning detection techniques. Section 3 presents the biological inspiration of the DCA, a summary of relevant developments in immunology, and rudimentary DC biology. This is followed by Sections 4 and 5, describing the abstraction process, a formalised description of the DCA and its implementation as an anomaly detector. This is followed by experimentation with its application as a port scan detector. Section 6 includes a sensitivity analysis of a selection of parameters. The paper concludes with a discussion of the results of the port scan investigation and suggestions for future work.

## 2. Anomaly Detection and Port Scanning

One notable application area of multi-sensor data fusion is anomaly detection, a technique used in Intrusion Detection, which uses behaviour based approaches to detect abuse and misuse of computer systems. Traditional approaches to computer security have relied on signature based approaches for the detection of intruders. Network based intrusion detection systems (IDS) such as Snort[25] cross reference patterns of network packets against a database of known intrusions. If a packet matches any of the signatures contained in the database an alert is generated, notifying the user of a potential intrusion. One problem with signature based approaches is that slightly modified intrusions or brand-new intrusions are not detected as they are not contained within the database resulting in false negatives.

Anomaly detection offers an alternative approach, by using a defined database of 'normal', either in terms of machine behaviour or user behaviour. Data at run time is compared against the normal profile and sufficient deviation causes the generation of alert. This is demonstrated through the research of the negative selection algorithm[13]which forms the majority of anomaly detection research within artificial immune systems. Unfortunately, defining what is normal is non-trivial and has a tendency to change over time, giving rise to systems with a high rate of false positives. To overcome the problems of false positives, a whole host of methods have been employed. This frequently involves adding a dynamic profiler to account for expected changes in the normal profile, or the use of more and disparate data sources. It is worthy of note that

In computer security, anomaly detection has been applied to a wide range of problems. This includes the detection of trojans, viruses, rootkits, network expoits, and distributed denial of service. As an application of anomaly detection in computer security, we examine the problem of detecting port-scans. They are a key tool in initiating an attack, and are frequently used in 'insider attacks' which are performed by authorised users.

### 2.1. General Principles of Port Scanning

Port Scanning is a technique of network cartography. It is used by system administrators to check specified hosts on their network for availability and to monitor services in use. However it can be subverted for more malicious purposes. Port scanning tools such as 'Network Mapper' (`nmap`) [23] can reveal information about hosts responding on a given set of network addresses. This information may be used by attackers to discover a set of target hosts which are operating services likely to be vulnerable to attack. It can also be used for an attacker to learn and understand the topology of a network in order to launch an attack such as a distributed denial of service.

A host on an IP network has one or more IP addresses. Each IP address has a range of $2^{16}$ TCP ports and $2^{16}$ UDP ports. Ports are simply a way of multiplexing many different types of communication through a single network address. This is why it is possible to download mail and surf the web at the same time. A program running on a network host may listen for requests on one or more (address, protocol, port) tuples. Many services typically listen for requests on standard port numbers (such as TCP port 80 for the HTTP service), though in reality, a service can be located on any port number. Port scanning involves probing a host to discover potential exploitable ports.

Instances of port scans differ from each other through a number of important properties. At a high level of abstraction there are two distinctions to be made. Firstly, modern networks are comprised of suites of various network protocols which offer different kinds of endpoints which can be useful to map. This work restricts discussion to IP networks in which there are IP address, TCP port and UDP port endpoints. Scans which map out these different types of endpoints use different methods. Secondly, attackers rarely wish to scan every possible endpoint reachable from their network, so a subset of endpoints are selected for mapping. The way in which the scope of the scan is restricted leads to a differing 'scan footprint'.

Once a scanner has created a list of endpoints, a probe is performed on each endpoint in order to obtain the scan results. As mentioned, various probe techniques are available depending on what kind of endpoint is being probed and for what information. The three main types of probe are:
  (i) Host probe: Determining if a given network address is assigned to a host
 (ii) Port probe: Determining if a service is listening at an (address, protocol, port) tuple
(iii) Service probe: Determining what kind of service is running over an (address, protocol, port) tuple

Host probes are typically carried out by sending ICMP echo requests to the IP address being queried. For this reason host scans are usually referred to as "ping scans" after the name of the UNIX program for sending these packets. If a host is associated with the queried IP address it may respond with an ICMP echo reply. However many systems simply do not respond to echo requests due to the potential for abuse. For this reason TCP probes may be sent to a port likely to be un-filtered (such as TCP port 80) and any response at all from that address is considered positive. If there are intervening routers between the scanner and the target host an ICMP host unreachable message may be generated for any traffic sent to an inactive address.

In the TCP/IP protocol suite, UDP and TCP port probes are possible. TCP port scans occur with much higher frequency than other types of scan. The simplest type of TCP probe connects to a port on a remote address and if the connection succeeds immediately closes the connection. A more stealthy approach, termed a "SYN scan" simply sends the first packet of the three-way handshake and uses the response packet to distinguish between open and closed ports. This usually requires super-user privileges. The only available technique for probing UDP ports is to send a packet containing random data to a UDP port on a remote host. If the port is open no response will be generated and if the port is closed an ICMP port unreachable error message is generated. Service scans are typically carried out after a port scan and lead to knowledge of the the type and version of operating system and network service software running on a remote host. In fact, the exact behaviour elicited by a host as a response to any of these probe types can be used in determining the operating system type and version.

These probing techniques may be combined with lists of endpoints to perform different types of scan. Three classes of scan footprint suffice to describe any particular scan type:
  (i) Horizontal Port Scan: Here the attacking host scans a range of IP addresses using the same port. This can reveal a set of 'live' hosts on a network with a specific open port. This is also used in several scanning worms based attacks. According to Staniford et al [27] this is the most common type of scan footprint.
 (ii) Vertical Port Scan: The attacking host sends several packets to the same IP address across a range of ports. This is used to target a specific host to examine any open ports or to uncover vulnerable running services. This can also be used to retrieve detailed information on the OS of the victim host.
(iii) Block Scan: This is a hybrid method combining a range of addresses with a range of ports. This is also used to target specific hosts. It is also used to generate 'hit-lists' for future attacks. Block scans can potentially take a very long time, hence the results of a ping scan may be used to dramatically reduce the number of endpoints to be probed.

Port scans may seem innocuous, but they can be used for malicious purposes. It is reasonable to use a port-scan as a model of an intrusion given that they frequently play some role in an attack, be it a targeted exploit or a scanning worm. Staniford et al are quoted as saying "... we detect in practice that almost all of them [unsolicited scans] come from compromised hosts and are very likely to be hostile". Suprisingly, port scan detection appears to be an under-researched area, while port scanning occurs frequently as a pre-cursor for more serious attacks. The detection of an ICMP ping scan forms the focus of the remainder of this section.

2.2. Port Scan Detection

Previous work in this area is suprisingly somewhat sparse. A number of IDS have the capability to detect some types of port scan[28,27,24] but most have so far relied on the assumption that X events occur within time frame Y. For example, Spice by Staniford et al [27] produces an alert every time a single IP connected to more than 15 hosts within time window Y. However, these types of technique cannot be used to detect stealthy scans which do not produce enough events within the specified time window.

The detection of scanning worms is a closely related and comparatively well researched area. Scanning worms frequently use port scans to generate a list of vulnerable hosts for propagation. Schecter et al [26] use a technique called reverse sequential hypothesis testing. This is based on connection analysis which determines the probability of a connection being anomalous. These data are combined with network flow information and the data sources correlated. Detection of scanning using the worm detection approach resulted in the detection of all but the stealthiest scans, namely those with a very slow scanning rate.

A worm detection technique pertinent to ICMP scan detection is the use of ICMP destination unreachable errors (Type 3 error) to detect the propagation of worms across a network. Bakos et al [2] used the capture and analysis of ICMP packets and packet flow to identify 'blooms' of ICMP traffic across a network. They assume that a high rate of Type 3 errors is indicitive of a worm. Early detection of scanning worms was achieved in the preliminary results presented. More details regarding the use of this technique in a realistic network scenario under more noisy test conditions have not been reported so far.

The idea of detecting the response to a scan as opposed to the scan itself is similar to the danger detection mechanisms which inspires the DCA. As opposed to examining incoming data to see if you are the recipient of a scan based attack, the outgoing data can be used to detect if your host is infected and is now scanning the local subnet. This approach is known as extrusion detection and has proven effective in the prevention of spam across a medium size network [4]. It has been shown that a high proportion of attacks, especially within a corporate setting, can originate from within the organisation itself as a result of misuse out of malice or ignorance. The detection of 'insider-attacks' is a pertinent problem, to which extrusion detection may prove useful.

2.3. Port Scanning Summary

Port scanning is both a useful tool for network administration and maliciously for use in the discovery of vulnerable hosts. Different types of scan are used for different purposes, with the most common type of scans based on the TCP protocol across a range of IP addresses, namely horizontal scans. ICMP ping scans are also popular as they are a very fast way of gaining network topology information, which can be used in future attacks. Detection techniques for port scans frequently rely on the assumption of detecting a number of events occuring within a time window. This is not effective in detecting more sophisticated scans. Scanning worm detection involves a number of the same principles. Alternative approaches include backward scan detection where the response of scanned hosts is used in place of detecting the port scan packets. 'Extrusion detection', where outgoing packets are examined, is useful for the detection of spam and could be used for the detection of insider attacks.

3. The Immune System: A DC's Perspective

The human immune system is a complex and robust system, viewed as a homeostastic protection agent[5]. It seeks out harmful pathogens, clearing them from the body and performing maintenance and repair. Classically the immune system is sub-divided into two distinct systems: the innate and adaptive immune system.

The innate immune system contains a variety of cells including macrophages and DCs amongst others, [15]. The innate immune system is the first line of defence against attack from invading organisms. The cells of the innate system express proteins on their surface, called receptors, and have the ability to detect and dispose of pathogens via ingestion through phagocytosis. The selectivity of the receptors for pathogenic material evolved within the development of our species and is passed down through the generations[7].

The adaptive immune system consists of two classes of cell, T-cells and B-cells. They differ from innate cells as their receptors are generated over the lifetime of the individual, not through the development of the species. The fine tuning of these receptors, performed during childhood, plays a key role in adaptation to previously un-encountered threats. For example, T-cells are selected in their early stage of development. Antigens, made of protein and derived from self cells, are presented to the naive T-cells. Those cells with a high affinity or can bind strongly to self antigen are deleted, leaving a set of detectors with receptors specifically designed to detect antigens which do not belong to the host. This forms the core of the self-nonself theory proposed in 1959 (described in [15]).

Since the 1970s immunology has developed in a number of significant ways. It was proposed that T-cell binding to pathogenic antigens is incapable of initiating immune activation without the presence of a second signal[15]. Investigation into vaccine development highlighted the need to add stimulatory molecules derived from pathogens (adjuvants) to innoculations in order for the process to be effective. Antigens in the innocculation have different structures than antigens belonging to self, yet an adverse response is not observed.

In addition to adjuvants, the immune system does not react to 'friendly' bacteria in the intestines, despite their prevalence. In the case of autoimmune diseases such as multiple sclerosis, the immune system reacts destructively against the body's own cells. Why should a system which has been filtered against self reactivity, respond actively to 'self' without any obvious cause? Self-nonself could not account for these imporant effects, so researchers turned their attentions to the cells of the innate immune system for answers.

In 1989, immunologist Charles Janeway and his colleugues proposed the infectious nonself model[14]. This is a two signal model that states that only antigens presented with co-stimulatory molecules (CSMs) can activate T-cells. T-cells do not reside in tissue, but are stored in lymph nodes, where they are given antigen by antigen presenting cells, which include DCs. Janeway showed that when DCs are exposed to 'signals', forming a class of molecules known as pathogen associated molecular patterns (PAMPs), matching T-cells became activated[14].

PAMPs are exogenous signals which are molecules produced exclusively by pathogens. The receipt of PAMPs is thought to enhance the binding between T-cell and DC. Foreign antigens are not recognised unless they are accompanied by PAMP signals which confirm their status as nonself. This explained why stimulatory adjuvants are necessary for immunisations to be successful. Unfortunately, the infectious nonself model can not explain the phenomena of autoimmunity.

The Danger Theory was proposed by controversial immunologist Polly Matzinger in 1994[18]. She stated that the immune system is controlled by the detection of damage to the body, not the detection of antigen structures or bacterial products. Matzinger proposes that signals do not come from exogenous sources, but are endogenous and produced by the cells of the tissue themselves. These endogenous signals are termed danger signals. The danger theory also proposes that the cells of the innate immune system can actively suppress an immune response in the absence of danger in the tissue. This is mediated through the recognition of 'tissue context' derived based on the balance between two types of cell death: necrosis and apoptosis[6].

Under healthy conditions, cells still die. Apoptosis or planned cell death regulates growth and development. During apoptosis the cell's internal contents are gracefully degraded. Genetic material is cut into orderly fragments and destructive enzymes known as lysosomes or 'suicide-sacs' digest the cell from the inside out. This prevents any loss of membrane integrity. Eventually the apoptosing cell shrinks and produces output signals e.g. tissue necrosis factor alpha.

DCs are sensitive to an increase in the signals of apoptosis and are attracted to the dying cell. Eventually the cell is found by a DC and ingested. Very little debris is left in the tissue and during this process. If the cell is ingested by an DC, the protiens contained within the cell are presented to the immune system as antigen in a 'safe' context, as the cell died of a normal process. The immune system is then tolerised to antigens with the same structure. This dynamic process is known as the mechanism of peripheral tolerance[20].

Not all cells die in this clean and controlled manner, as in the case of cell death as a result of injury. Cell stress can occur through irradiation, shock, hypoxia or pathogenic infection, leading to the death of the cell via necrosis. Unlike apoptosis the internal cell contents degrade chaotically and the cell membrane loses its integrity. Irregular fragments of DNA are produced and oxidised to become uric acid crystals, with heat shock proteins and other hydrophobic compounds released from the cell. These molecules were previously separated from the tissue fluid by the cell membrane and form the danger signals[20]. Dendritic cells are sensitive to changes in concentration of the molecules released as a result of necrosis. Upon the detection of danger, the DC migrates from the tissue and presents any collected debris as antigen to T-cells, causing activation.

To summarise, danger theory states that the immune system is activated by DCs upon receipt of danger signals. DCs have the ability to combine signals from apoptosis, necrosis and PAMPs, and to use this information to instruct the immune system to respond appropriately. Unlike the assertions of self-nonself, this model emphasises that signals from the environment dictate the behaviour of the immune system, not the structure of antigens. Dendritic cells are the natural data fusion agents which have the ability to combine both endogenous and exogenous signals with antigen to detect invading pathogens and to maintain tolerance[22].

3.1. Introducing Dendritic Cells

To derive an algorithm based on the danger theory it is necessary to understand the mechanisms used by DCs to detect pathogens. DCs belong to a family of cells known as macrophages, whose function is to clear the tissue of debris. Unlike other macrophages, DCs have a specialised role as professional antigen presenting cells and control the activation state of T-cells in the lymph nodes. The term "dendritic cell" refers to the fact that they can have long finger like projections which resemble dendrites. They are a class of cell, which exist in three distinct states: immature (iDC), semi-mature(smDC) and mature (mDC), shown in Figure 1. The state of differentiation is determined by the relative signal concentration they receive while in the tissue.

**Immature DC**  **'Semi-mature' DC**  **Mature DC**

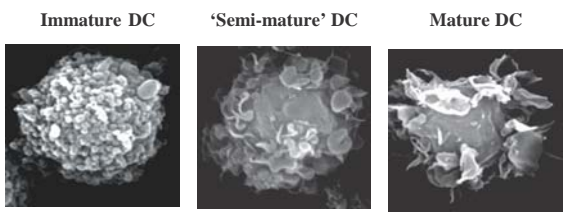

Fig. 1. Immature , 'Semi-mature' and Mature Dendritic Cells ESMicrograph picture (see acknowlegements)

3.2. Immature DCs (iDC)

On arrival in the tissue, DCs are found in an immature state[17]. Here, iDCs collect debris, some of which is used as antigen. Antigens are complexed with an auxilliary molecule necessary for T-cell binding and are transported to the cell membrane for presentation. In addition to antigen processing, DCs can sense the various signals that may be present in the tissue through receptors expressed on the cell's surface. These receptors are sensitive to PAMPs, danger signals and 'safe signals'. The relative proportions and potency of the different signals determines the iDC's terminal state of differentiation. Receipt of signals causes changes to the function, morphology and behaviour of the iDC. The result of exposure to signals causes the production of molecules called cytokines which can either activate or suppress the immune system. It is important to note that iDCs cannot present antigen directly to or activate T-cells directly as they do not produce the necessary cytokines.

3.3. Mature DCs (mDC)

DCs which have the ability to both present antigen and activate T-cells are termed mature DCs. For an iDC to become an mDC, the iDC must be exposed to a greater quantity of either PAMPs or danger signals than safe signals. Exposure to signals takes place during the iDCs antigen collection stage. Sufficient exposure to PAMPs and danger signals causes the DC to cease antigen collection and migrate from the tissue to the lymph node. The high concentration of T-cells in the lymph nodes increases the probability of a successful antigen match between DC and T-cell. During the migration, the iDC changes morphologically to become a mDC by developing whispy finger-like projections which gives them an increased surface area. This further increases the probabiltiy of binding with a T-cell[22]. An increase in surface area makes the mDC more suitable for antigen presentation rather than collection.

Most importantly, mDCs produce an inflammatory cytokine called Interleukin-12, which stimulates T-cell activation. Additionally the mDC produces costimulatory molecules (CSMs), which are known to facilitate the antigen presenting process [21]. PAMPs and danger signals detected in the tissue while in the immature phase are thought to be responsible for the production of Interleukin-12 and CSMs.

3.4. Semi-mature DCs (smDC)

Under apoptotic conditions, exposure to safe signals diverts the iDC to a terminal state known as 'semi-mature'. They appear morphologically very similar to mDCs and can present antigen, yet they do not have the ability to activate T-cells. Instead of secreting Interleukin-12, the smDC produces Interleukin-10.

Interleukin-10 suppresses T-cells which match the presented antigen. Antigens collected with safe signals are presented in a tolerogenic context and lead to unresponsiveness to those antigens. Evidence suggests that safe signals have a greater influence on DCs than PAMPs and danger signals, and can actively inhibit the production of Interleukin-12 while increasing production of Interleukin-10[32]. This is a natural mechanism designed to stop the immune system over reacting to antigens. In essence, the immune system expends considerable time and energy preventing reactions to harmless antigen or self antigen.

3.5. Summary

Dendritic cells are antigen presenting cells which have the power to control the adaptive immune responce. DCs initial function is to collect debris from the tissue called antigen. Instructions to the adaptive immune system are derived based on the relative signal concentration found in the tissue where immature DCs reside, represented in Figure 2. Three signal categories have been discovered. Pathogenically derived PAMPs and danger signals from dying cells cause the DC to mature and present antigen to the effector T-cells. Conversely, signals collected as a result of apoptotic death cause the DC to mature to a different 'semi-mature' state. The smDCs cannot activate T- cells, but cause presentation of antigens in a tolerogenic context, vital to the prevention of autoimmunity. The mechanism by which DCs process signals is intricate, and the three signal concentrations are fused within the cell to influence the resulting output of CSMs, Interleukin-10 and Interleukin-12. This output informs the immune system how to respond appropriately.

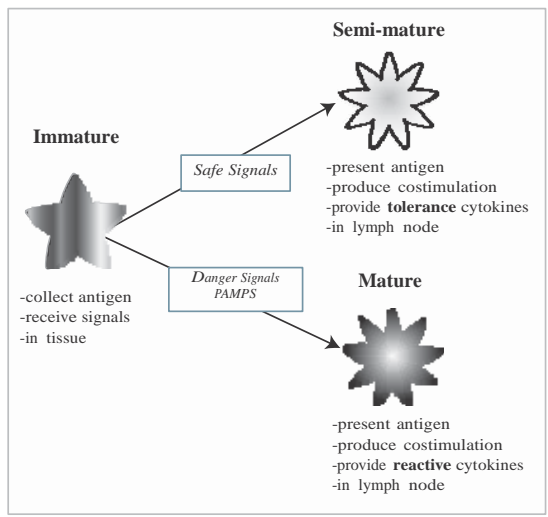

Fig. 2. An abstract view of DC maturation and signals re- quired for differentiation. CKs denote cytokines.

4. From in vivo to in silico

Through close collaboration with immunologists [32], we have abstracted what we believe to be the essential features of DC biology. DCs are examined from a cellular perspective, which includes the differentiation states, interaction with signals and antigen. Representations of signals, antigen and the different DC states form the core of this abstraction. The following properties of DC function are used, and summarised in Figure 2:
- Signals and Antigen:
(i) Exposure to signals initiates maturity of an iDC not the collection of antigen.
(ii) The quantity of output signals produced is determined by processing input signals from the environment, and can be viewed as an interpretation of the relative input signal strength.

(iii) Input signals to a DC are either PAMPs derived from pathogenic signatures, danger signals from damaged tissue or safe signals from normal cell death.

(iv) Overall decision of tolerance or activation is dictated by the combined behaviour of a population of DCs.

- Immature DCs:
(i) iDCs can differentiate to become either mDCs or smDCs.
(ii) The path of differentiation is dictated by the complement of signals to which an iDC is exposed.
(iii) Each iDC can sample multiple antigens, which are internalised and re-presented with cytokines reflecting the context.

- Semi-Mature DCs:
(i) Safe signals suppress the production of the mature output signal.
(ii) The smDCs produce a different output signal which confirms that the presented antigen was collected in a normal environment.

- Mature DCs:
(i) Both mDCs and smDCs can present antigen by producing costimulatory molecules.
(ii) The mDCs produce an output signal which confirms that the presented antigen were collected in a context of danger and damage

## 5. The Dendritic Cell Algorithm

The DCA is an algorithm which uses a population of agent-like, software-based artificial DCs which combine data from disparate sources. This description of the DCA is based on an implemented version of the algorithm made possible through the use of the libtissue framework[31].

### 5.1. Libtissue

The Danger Project [1] has produced a variety of research outcomes alongside the DCA: the development of danger theory and DC based immunology[32]; a framework for developing immune inspired algorithms called libtissue[31]; an investigation into the interactions between the innate and adaptive immune system; artificial tissue [3] and the application of a naive version of the DCA for the security of sensor networks. Libtissue is the API used within the Danger Project for the testing of ideas and algorithms, as shown in the works of Twycross [29] and Greensmith et al [10].

Libtissue is a library implemented in C which assists the implementation and testing of immune inspired algorithms on real-world data. It is based on principles of innate immunology[30] [31], through the use of compartmentalisation, and uses techniques from modeling, simulation and artificial-life. It allows researchers to implement algorithms as a collection of cells, antigen and signals interacting within a specified compartment. The implementation has a client/server architecture, with communication perfromed via sockets using the SCTP protocol. This architecture separates data collection using clients, from data processing on a server, as shown in Figure 3.

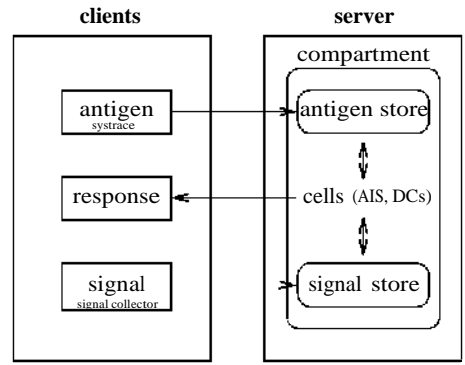

Fig. 3. Architecture used to support the DCA. Input data are processes via signal and antigen clients. The algorithm utilises this data and resides on a server.

Input data are processed using libtissue clients, which transform raw data into antigen and signals. Algorithms can be implemented within the libtissue server, as libtissue provides a convenient programming environment. Antigen and signal sources can be added to libtissue servers, facilitating the testing of the same algorithm with a number of different data sources. Input data from the client passed to and represented in a compartment contained on a server known as a tissue compartment. This is a space in which cells, signals and antigen interact. Each tissue compartment has a fixed-size antigen store where antigen provided by libtissue clients is placed. The tissue compartment also stores levels of signals, set either by the input clients or cells.

## 5.2. Abstract View of the DCA

The DCA is implemented as a libtissue tissue server. Input signals are combined with a second source of data, such as a data item ID, or program ID number. This is achieved through using a population of artificial DCs to perform aggregate sampling and data processing. Using multiple DCs means that multiple data items in the form of antigen are sampled multiple times. If a single DC presents incorrect information, it becomes inconsequential provided that the majority of DCs derive the correct context. The sampling of data is combined with context information received during the antigen collection process.

Different combinations of input signals result in two different antigen contexts. Semi-mature antigen context implies antigen data was collected under normal conditions, whereas a mature antigen context signifies a potentially anomalous data item. The nature of the response is determined by measuring the number of DCs that are fully mature, represented by a value, MCAV - the mature context antigen value. If the DCA functions as intended, the closer this value is to 1, the greater the probability that the antigen is anomalous. The MCAV is used to assess the degree of anomaly of a given antigen. By applying thresholds at various levels, analysis can be performed to assess the anomaly detection capabilities of the algorithm.

The DCA has three stages: initialisation, update and aggregation. Initialisation involves setting various parameters and is followed by the update stage. The update stage can be decomposed into tissue update and cell cycle. Both the tissue update and cell cycle form the `libtissue` tissue server. Signal data are fed from the data-source to the tissue server through the tissue client. A graphical representation of this process can be seen in Figure 4.

The tissue update is a continuous process, whereby the values of the tissue data structures are refreshed. This occurs on an event-driven basis, with values for signals and antigen updated each time new data appears in the system. Antigen data enters tissue update in the same, event driven manner. The updated signals provide the input signals for the population of DCs.

The cell cycle is a discrete process occurring at a user defined rate. In this paper, one cell cycle is performed per second. Signal and antigen from the tissue data structures are accessed by the DCs during the cell cycle. This includes an update of every DC in the system with new signal values and antigen. The cell cycle and update of tissue continues until a stopping criteria is reached, usually until all antigen are processed. Finally, the aggregation stage is initiated, where all collected antigen are subsequently analysed and the MCAV per antigen derived.

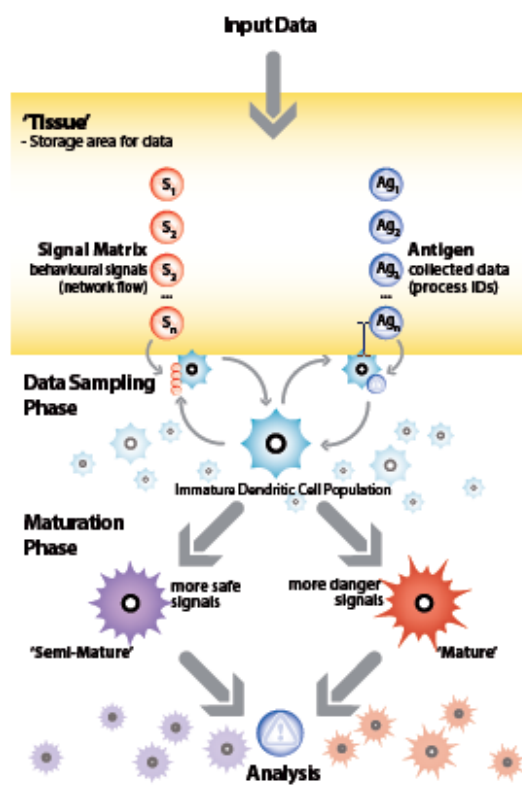

Fig. 4. Illustration of the DCA showing data input, continuous sampling, the maturation process and aggregate analysis.

## 5.3. Parameters and Structures

The algorithm is described using the following terms.
- Indices:
  $i = 0, ..., I$ input signal index;
  $j = 0, ..., J$ input signal category index;
  $k = 0, ..., K$ tissue antigen index;
  $l = 0, ..., L$ DC cycle index;
  $m = 0, ..., M$ DC index;
  $n = 0, ..., N$ DC antigen index;
  $p = 0, ..., P$ DC output signal index.

– Parameters:
  I = number of input signals per category;
  J = number of categories of input signal;
  K = number of antigen in tissue antigen vector;
  L = number of DC cycles;
  M = number of DCs in population;
  N = DC antigen vector size ;
  P = number of output signals per DC;
  Q = number of antigens actually sampled per DC for one cycle;
  R = maximum number of antigen collected per DC for one cycle (DC antigen receptors) ;
  $T_{max}$ = tissue antigen vector size.

– Data Structures:
  $DC_m = \{s(m), a(m), \bar{o}_p(m), t(m)\}$- a DC within the population;
  T = {S, A} - the tissue;
  S = tissue signal matrix;
  $s_{ij}$ = a signal type i, category j in the signal matrix S;
  A = tissue antigen vector;
  $a_k$ = antigen k in the tissue antigen vector;
  s(m) = signal matrix of DC (m);
  a(m) = antigen vector of of DC (m);
  $o_p(m)$ = output signal p of o DC (m);
  $\bar{o}_p(m)$ = cumulative output signal p of $DC_m$;
  $t(m)$ = migration threshold of $DC_m$;
  $w_{ijp}$ = transforming weight from $s_{ij}$ $o_p$.

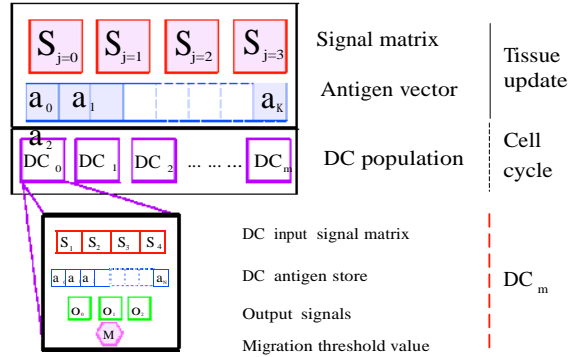

Fig. 5. Tissue and Cell Update components, where $S_{i,j}$ is reduced to $S_j$.

The data structures are represented graphically in Figure 5. Each $DC_m$ transforms each value of s(m) to $o_p(m)$ using the following equation with suggested values for weightings given in Table 1 and presented in Figure 6. Both the equation and weights are derived from observing experiments [32] performed on natural DCs for the purpose of their relative derivation. In the DCA each component of the antigen vector provides the capacity for storage of individual antigen. Although each DC samples the same input antigen vector, each DC samples a different component, potentially containing an antigen. Each DC samples the same input signal matrix and each 'component of the signal matrix.

$$o_p(m) = \frac{\sum_i \sum_{j \neq 3} W_{ijp} s_{ij}(m)}{\sum_i \sum_{j \neq 3} |W_{ijp}|} \quad \forall p$$

Table 1
Weights used for signal processing

| $w_{ijp}$ | j = 0 | j = 1 | j = 2 |
|---|---|---|---|
| p = 1 | 2 | 1 | 2 |
| p = 2 | 0 | 0 | 3 |
| p = 3 | 2 | 1 | -3 |

The tissue has containers for signal and antigen values, namely S and A. In the current implementation of the DCA, there are three categories of signal (j = 2) and 1 signal per category (i = 0). The categories are derived from the three signal model of DC behaviour described in Section 2 where: $s_{0,0}$ = PAMP signals, $s_{0,1}$ = danger signals, and $s_{0,2}$ = safe signals. An antigen store is constructed for use within the tissue cycle where all DCs in the population collect antigen, which is also introduced to the tissue in an event driven manner.

The cell cycle maintains all DC data structures. This includes the maintenance of a population of DCs, which form a sampling set of size M. Each DC has an input signal matrix, antigen vector, output signals, and migration threshold. The internal values of $DC_m$ are updated, based on current data in the tissue signal matrix and antigen vector. The DC input signals, s(m) use the identical mapping for signal categories as tissue s and are updated every cell cycle iteration. Each s(m) for $DC_m$ is updated via an overwrite every cell cycle. These values are used to calculate output signal values, $o_{p(m)}$, for $DC_m$, which are added cumulatively over a number of cell cycles to form $\bar{o}_p(m)$, where p = 0 is costimulatory value, p = 1 is the mature DC output signal, and p = 2 is the semi-mature DC output signal. With each cell update, DCs sample R antigens from the tissue antigen vector A.

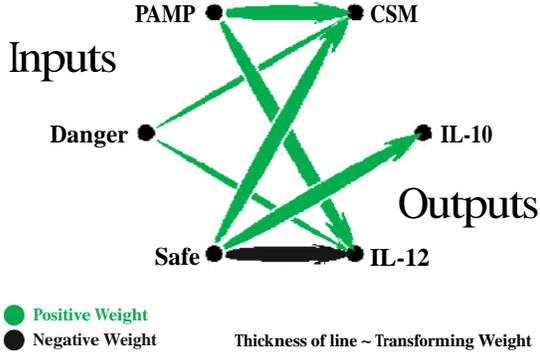

Fig. 6. A representation of the three calculations performed by each DC per update cycle, to derive the cells outputs through fusing together the signal inputs.

### 5.4. The DCA

The following pseudocode shows the initialisation stage, cycle stage, tissue update and cell cycle.

```
initialise parameters
{I, J, K, L, M, N, O, P, Q}
while (l< L)
   update A and S
   for m = 0 to M
      for 0 to Q
       DC_m samples Q antigen from A
      for all i = 0 to I and all j = 0 to J
       s_ij^DC = s_ij
      for n = 0 to N
       DC_m processes a_nm^DC
      for p to P
       compute o_p
       ō_p(m) = ō_p(m) + o_p
      if o_0(m) > t_m
       DC_m removed from population
       DC_m migrate, print antigen and
context
       DC_m reset antigen vector and all
signals
   l++
analyse antigen and calculate MCAV
```

### 5.5. Antigen Aggregation

Once $DC_m$ has been removed from the population, the contents of a(m) and values $\bar{o}_p(m)$ are logged to a file for the aggregation stage. Once completed, s(m), a(m) and $\bar{o}_p(m)$ are all reset, and $DC_m$ is returned to the sampling population. The re-cycling of DCs continues until the stopping condition is met (l = L). Once all data has been processed by the DCs, the output log of antigen-plus-context is analysed. The same antigen is presented multiple time with different context values. This information is recorded in a log file. The total fraction of mature DCs presenting said antigen (where $\bar{o}_1 > \bar{o}_2$) is divided by the total amount of times the antigen was presented namely $\bar{o}_1/(\bar{o}_1 + \bar{o}_2)$. This is used to calculate the mean mature context antigen value or MCAV.

### 5.6. Signals and Antigen

An integral part of DC function is the ability to combine multiple signals to influence the behaviour of the cells. The different input signals have different effects on cell behaviour as described in Section 3. The semantics of the different category of signal are derived from the study of the influence of the different signals on DCs in vitro. Definitions of the characteristics of each signal category are given below, with an example of an actual signal per category. This categorisation forms the signal selection schema.

– PAMP - $s_{i0}$ e.g. the number of error messages generated per second by a failed network connection
 (i) a signature of abnormal behaviour e.g. an error message
 (ii) a high degree of confidence of abnormality associated with an increase in this signal strength
– Danger signal - $s_{i1}$ e.g. the number of transmitted network packets per second
 (i) measure of an attribute which significantly increases in response to abnormal behaviour
 (ii) a moderate degree of confidence of abnormality with increased level of this signal, though at a low signal strength can represent normal behaviour.
– Safe signal - $s_{i2}$ E.g. the inverse rate of change of number of network packets per second. A high rate of change equals a low safe signal level and vice versa.
 (i) a confident indicator of normal behaviour in a

predictable manner or a measure of steady-behaviour

(ii) measure of an attribute which increases signal concentration due to the lack of change in strength

Signals, though interesting, are inconsequential without antigen. To a DC, antigen is an element which is carried and presented to a T-cell, without regard for the structure of the antigen. Antigen is the data to be classified, and works well in the form of an identifier, be it an anomalous process ID[10] or the ID of a data item [9]. At this stage, minimal antigen processing is performed and the antigen presented is an identical copy of the antigen collected. Detection is performed through the correlation of antigen with fused signals. By processing of antigen, this refers to the process by which antigen is collected and presented for analysis by the DCs - it is noteworthy that no changes are made to the actual value of the antigen, it is sampled whole.

The DCA could be interpreted as a neural network if its goal was to purely classify based on weighed sums alone. However the algorithm is not designed for the purpose of classification, but sorts input data, in the form of antigen, through the use of data-fused signals. The signals are aggregated through time and across a population of cells, which is different to the processing performed by a series of neural networks.

## 6. PSI: Ping Scan Investigation

The purpose of this investigation is as follows:
(i) To apply the DCA to anomaly detection through bio-inspired data fusion.
(ii) To show how the system responds to the modification of signal mappings.
(iii) To understand the sensitivity of the system parameters and the sensitivity of the weights of the signal processing function.

In this paper, port scanning is used as a model intrusion, and is described in Section 2. The DCA is applied to the detection of an outgoing port scan across a range of IP addresses, based on the ICMP 'ping' protocol. It is assumed that it is possible for the attacker to gain access to the machine using a password cracking utility.

The type of scan used in this investigation is an `nmap` ICMP 'ping' scan. This type of scan is particularly suitable for the purpose of these experiments as it is suitable for use on a network as it uses minimal network resources and is a short duration scan (10-30 seconds in duration). Ping scans involve a victim machine, connected to a medium sized subnet of 100-200 machines, which has been subverted by our hypothetical intruder. The premise is that the intruder has logged into the victim machine remotely via ssh and aims to retreive a list of hosts running within a similar IP address range. During the scan, the victim machine sends ICMP probes to other hosts, specified at run time. The nmap scan program reports back on the status of the scanned addresses as either appearing to be down or up. This allows an attacker to generate a list of hosts currently running within a range of IP addresses. This scan does not require root privileges and is one of the fastest scans available. Ping scans also retrieve DNS information, resolving the IP address of available hosts.

### 6.1. Data Sessions

Two types of data session are used in this investigation. An attack session consists of a ping scan embedded within a 70 second ssh session. Four processes (running programs) of interest are identified in these sessions for the purpose of analysis including: ssh demon; bash shell; nmap scan program; and the pts sshd process which is the parent of the nmap scan. The ssh demon and the bash shell are normal process which occur in the attack session. The scan uses a range of 1020 IP addresses across a class C network. The normal session also involves a remote log in via ssh, and also contains the transfer of a file from the victim machine to a remote server, via scp. Again, four processes of interest are identified: bash shell; sshd; x-forwarding agent; secure copy of a 2.5MB file. Ten datasets are generated for both the attack and normal protocols.

### 6.2. Signals

Three signal categories are used, with one signal per category throughout this investigation. The signals used are defined in Section 3, where they are placed in context with their biological inspiration. Signals are collected from kernel statistics using bash scripting, and are processed, normalised and combined with antigen to form a log file. All signals are normalised real-values within a range of 0-100 for the PAMP and danger signal and 0-10 for the safe signal. It is important to note that preliminary examinations of the input signal data indicate that analysis of the signals individually is insufficient to indicate anomalies [11], which is further highlighted in the DCLite experiment.

The PAMP signal is generated from the rate of ICMP destination unreachable errors recieved per second. When a ping scan is used, ICMP packets are sent to the machines specified. Frequently the range of machines is specified as a block, for example X X.X X.20.1-254 would scan all addresses on the '.20' subnet [1] . It is likely that numerous machines within that range will not accept ping probes and hence a DU error is sent back to the scanning machine, as a signature of suspicious activity.

The danger signal is derived from the number of sent network packets per second. An increase in network traffic sent from the machine can be an indicator of anomalous behaviour. However, under certian circumstances, such as uploading files via a torrent client or over peer-to-peer networks, this is not as useful.

To complement this signal, the safe signal is derived using the first order derivative of packets/s, namely the rate of change of packets/s. This is based on the assumption that anomalous traffic produces 'bursty' rates of sending, whereas uploading large files will not have such a variable rate of change. To derive the safe signal, a maximum value for rate of change is defined. The more variable the rate of change, the greater the decrement of the maximum value. This is the inverse rate of change of packets per second. This counters an increased danger signal value under 'normal' conditions, and may assist in reducing false positives. A sketch of the input signals for both sessions is represented in Figure 7, where 0-50 seconds shows signals during a ping scan and 51-75 seconds shows the normal file transfer.

### 6.3. Antigen

In these experiments the signals are used to detect the anomalous nmap process and its sshd parent in the attack scenario, and actively prevent a response to the scp file transfer. This cannot be acheived by signals alone, as antigen is required to correlate the signals to the active culprit processes. During each session, all processes spawned by the controlling ssh sessions are monitored using the `strace` tool. Each of the processes is assigned a number identifier (PID) by the operating system. To run, each process invokes the use of system calls. The more active the process, the more system calls it makes. As antigen, each system call is captured and converted into an antigen, with a value of the PID to which the system call belongs. In a similar manner to the signals, output of this process is logged and combined with the signal data to form the datasets for these experiments. The multiplicity of input antigen facilitates the function of the algorithm, encompassing a DCs ability to collect and process multiple antigen fragments of identical structure.

---

[1] Full IP addresses not given, adhering with our organisa- tion's security policy

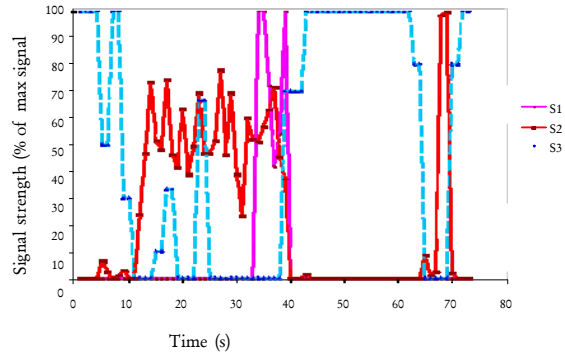

Fig. 7. A sketch of the input signals for both attack and normal sessions, where the left hand side of the figure rep- resents the attack dataset, and the right normal.

### 6.4. Experiments

As shown in Section 5, the algorithm implemented with libtissue has numerous parameters. It is necessary to understand the effect on the system by changing these parameters in order to understand the behaviour of the DCA. The experiments performed assist in clarifying these effect, and fall into three convenient categories:

Series-1: Investigate signal mappings: does incorrect data mapping influence the detection rates?
Series-2: Sensitivity of libtissue related parameters: which parameters can influence the system and at what value?
Series-3: Sensitivity of the weights of the signal processing equation: how to these weights relate to each other and what effect on detection arises from variation in the values?

## 6.5. Series-1

The aim of this series is to change the input signal mappings to assess the validity of this implementation. The mapping of the input signal category to the raw data attributes is controlled primarily by the weights of the signal processing equation. By performing experiments such as switching the PAMP and safe signal we predict that the system would respond with a very high rate of false positives. This information is used to validate the use of this algorithm on this particular problem. Three input signals are used in series-1 inclusive of one PAMP signal ($s_{0,0}$), one danger signal ($s_{0,1}$,) and one safe signal ($s_{0,2}$). In order to understand the principles of mapping signals to categories each chosen input for the signals is used per category.

The permutations of this experiment are shown in Table 2. We hypothesise that the DCA will not lose detection accuracy when the incorrect mapping involves the PAMP and danger signal, as these signals affect the DCs in a similar manner. Conversely a mapping reversal between the danger and safe signal may result in a poor performance as they are treated differently in the signal processing function. All experiments in this series are tested using all 20 attack and normal datasets, with three repeats of each run per dataset. Similar experiments involving combinations of 2, 3 and 4 input signals are presented in [11], to which the interested reader should refer.

Table 2
Experiment codes and signal mappings

| Experiement Code | $S_{0,0}$ | $S_{0,1}$ | $S_{0,2}$ |
|---|---|---|---|
| M1 | P | D | S |
| M2 | D | P | S |
| M3 | S | D | P |
| M4 | P | S | D |
| M5 | S | P | D |
| M6 | D | S | P |

## 6.6. Series-2

Numerous parameters within `libtissue` are used to define the behaviour of the artificial DC and the properties of the compartments. As a result several values which may influence the system need to be investigated to assist in understanding the algorithm. A summary of the series-2 experiment is presented in Table 3. The ten attack datasets are used for this series, providing examples of both normal and anomalous data, with three repeats of each run per dataset. Four key libtissue parameters are investigated:

(i) Number of DCs created (C)
(ii) DC antigen vector size (N)
(iii) Number of DC antigen receptors (R)
(iv) Size of tissue antigen vector (Tmax)

Table 3
Experiement codes and default parameter settings

| Experiment Code | Parameter Values |
|---|---|
| C | 10; 100; 200; 500 |
| N | 1; 2; 5; 10; 25; 50; 100 |
| R | 1; 2; 5; 10; 20 |
| Tmax | 50; 500; 1000; 5000; 10000 |

## 6.7. Series-3

Essentially, each DC in the sampling population performs data fusion through combining multiple signals from disparate sources to produce output signals, which are then correlated with data in the form of antigen. The combination of the input signals is achieved using the signal processing equation described in Section 5, where processing is performed three times on the the input signals to produce three different output signals. Initially the weights chosen for this purpose were derived from empirical biological data. Indeed, the inter-relationship between the weights (as shown in Table 4) remains inspired by these data, with all weights deriving from the weight of the PAMP signals. Two weights are investigated, W1 and W2.

Preliminary tests and prior experience with the DCA indicate values for W1 and W2 should lie within a range of 0 and 20 if the maturation threshold is 60 (+/- 50%). An exhaustive search of the following values is performed: 0.5; 1; 2; 4; 8; and 16. This results in a total of 36 experiments. One attack dataset is selected at random for use with three runs performed per parameter combination.

## 6.8. Parameters and Settings

All experiments are performed on an AMD Athlon 1GHz Debian linux machine (kernel 2.4.10). The algorithm is implemented within the `libtissue` framework, implemented in C (gcc 4.0.2) with interprocess communication facilitated by the SCTP protocol. All signals are derived using signal collection scripts, with values taken from the 'proc' filesystem. Unless stated otherwise, default parameters for all experiments are presented in Table 5.

Table 4
Derivation and interrelationship between weights in the signal processing equation

| Output Signal | Input Signal | Weight |
|---|---|---|
| $o_0$ | $s_{0,0}$ | W 1 |
|  | $s_{0,1}$ | W 1/2 |
|  | $s_{0,2}$ | W 1 * 1.5 |
| $o_1$ | $s_{0,0}$ | 0 |
|  | $s_{0,1}$ | 0 |
|  | $s_{0,2}$ | 1 |
| $o_2$ | $s_{0,0}$ | W 2 |
|  | $s_{0,1}$ | W 2/2 |
|  | $s_{0,2}$ | W 2 * -1.5 |

Table 5
Default parameter settings

| Name | Symbol | Value |
|---|---|---|
| Number of signals per category | I | 1 |
| Number of signals categories | J | 3 |
| Max number of tissue antigen | K | 500 |
| Number of cells | M | 100 |
| Max number of antigen per DC | N | 50 |
| Number of output signals per DC | P | 3 |
| Number of DC antigen receptors | Q | 1 |

6.9. Results

In this section results for all experimental series are presented, showing that the DCA can be used as a ping based port scan detection system. In all experiments the MCAV coefficient is employed to assess the behaviour and function of the DCA. The MCAV is the mature context antigen value and is a number between zero and one. The closer this value is to one, the higher the probability that the monitored process is anomalous as more antigen are presented in by the algorithm in the mature context i.e. the anomalous context. Each type of antigen is given a MCAV coefficient value which can be compared against a threshold. Once a threshold is applied, it is then possible to classify antigen as either 'normal' or 'anomalous' and therefore the relevant rates of true and false positives can be shown. This calculation is used throughout this section. The results of each series of experiments are presented with the relevant statistics shown.

6.9.1. Series-1

The graph presented in Figure 8 represents a summary of results for the attack datasets used with the original and correct mapping (M1). MCAVs generated by the two anomalous processes are 0.82 (with a standard deviation, stdev, of 0.11) for the nmap process and 0.67 (stdev.= 0.22) for the parent pts process. All statistical tests are performed using a paired t-test, where p = 0.05, used whenever 'statistical significance' is stated. These values are statistically significantly higher than the MCAVs produced for the normal processes. The bash MCAV is 0.02 (stdev=0.04) and the sshd MCAV is 0.18 (stdev=0.24). The variance (stdev divided by MCAV) values are also larger for the two normal processes in contrast the the anomalous items.

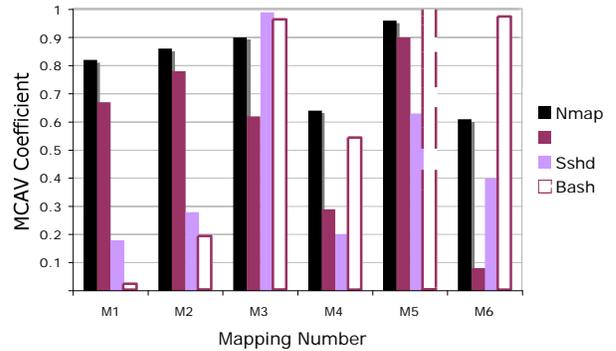

Fig. 8. MCAV values for all processes and all mappings for the attack datasets. Ten attack datasets are used, each point representing a mean of 30 values, as number of runs per dataset =3.

Figure 8 shows the MCAVs generated per process for each mapping, across the 10 attack datasets. The intended mapping (M1) is used as a baseline to which all other mappings are compared. The results for M1 and M2 are similar, with high MCAVs for the anomalous processes and low values for the normal items. Statistically, significant differences are shown between the MCAVs the bash process, as the bash MCAV in M1 is 0.02 as opposed to 0.27 for M2. This implies that incorrect mapping between PAMP and danger signals would not impair detection, save for a slight increase in the rate of false positives. M3 produced significantly higher MCAV for the two normal processes than M1, yet only minor differences in the detection of the anomalous processes. Incorrect mapping of safe signals as PAMPs leads to an increased rate of false positives.

In M4 the MCAVs for the anomalous processes are significantly smaller ($p > 0.05$) and significantly larger for the normal processes when compared to in M4 is 0.64, which is significantly lower than the 0.82 reached using M1. This trend is also shown in M6, with a nmap MCAV of 0.61 (stdev=0.37). M5 exhibited a similar increase in the MCAV of the normal processes, yet interestingly produced the highest MCAV for the anomalous processes, with the lowest standard deviation for the detection of the nmap process. However, normal MCAV values are significantly higher than observed in M1. All MCAVs for this experiment are shown with their standard deviations in Table 6.

Table 6
MCAV values for each experiment across each dataset.

| Expt. | Attack | | | | | | | |
|---|---|---|---|---|---|---|---|---|
| | nmap | | pts | | bash | | sshd | |
| | mean | stdev | mean | stdev | mean | stdev | mean | stdev |
| M1 | 0.82 | 0.04 | 0.67 | 0.11 | 0.18 | 0.22 | 0.02 | 0.24 |
| M2 | 0.86 | 0.27 | 0.78 | 0.12 | 0.28 | 0.27 | 0.19 | 0.35 |
| M3 | 0.90 | 0.04 | 0.62 | 0.13 | 0.99 | 0.33 | 0.96 | 0.02 |
| M4 | 0.64 | 0.29 | 0.29 | 0.29 | 0.20 | 0.28 | 0.54 | 0.05 |
| M5 | 0.96 | 0.03 | 0.90 | 0.10 | 0.63 | 0.32 | 1.00 | 0.00 |
| M6 | 0.61 | 0.37 | 0.08 | 0.06 | 0.40 | 0.21 | 0.97 | 0.05 |

For the normal datasets, similar trends are evident, as shown in Figure 9. M1 shows very low MCAVs for all processes, indicating a low rate of false positives. M2 is similar, also producing low values for all processes of interest. The scp process produced a higher MCAV of 0.22 in M2, though this is not statistically significant. M3 produced the maximum MCAV of 1 for all processes, while M4 produced values over 0.5. Experiment M5 produced interesting results. Despite poor performance for the attack dataset, high values of MCAV are not present in the normal dataset. The MCAV for the M5 scp is 0.13, which is not significantly different to the results found for M1. M6 produces similar results to M3, but the MCAVs are not as high (statistically significant, $p > 0.05$).

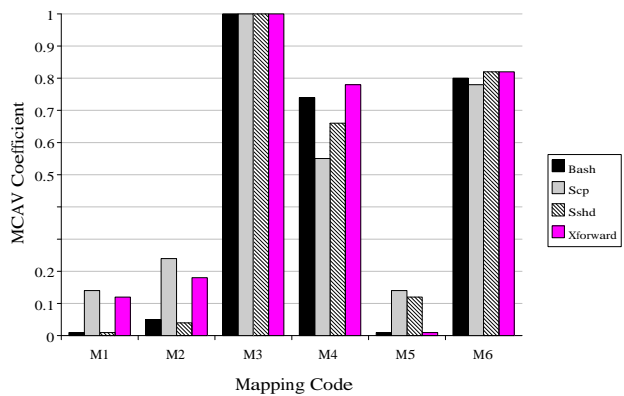

Fig. 9. Response varied signal mappings for the normal sessions. Each processes of interest is represented individually. Values represent mean MCAV coefficients from ten datasets, where number of runs = 3.

6.9.2. Series-2

In this series, various DCA parameters are assessed. Cell numbers parameter results are presented in Figure 10. This shows MCAVs per process, and each experiment is represented within each process. A high value of MCAV is shown for the nmap process for all values of cell number above 100. Where the number of cells is equal to 10, the MCAV is greatly reduced, from 0.9 to 0.1. The standard deviations of these values increases from 0.3 in C10, to 0.1 for all other cell numbers. Similarly no significant difference was found between C100, C200 and C500 for any of the processes. C10 also produced higher MCAVs for the normal processes than for any other cell number.

The DC antigen vector size parameter results are summarised in Figure 11. No significant differences are found when this parameter is varied, for any of the processes of interest. Marginally impaired performance is shown when this size is set to 100, but this was shown to not be significant. One explanation for this parameter's insensitivity is that the number of antigen is less influential than the signals they are collected. Further analysis is performed using the number of antigens processed to understand exactly the reasons for this effect.

Hence, investigations in to the number of antigen sampled by one DC per iteration (number of antigen receptors) is performed, and the results are presented in Figure 12. This shows no significant difference in the MCAV values for the detection of the two anomalous processes. Data regarding the MCAVs of the normal processes suggests that an increased number of antigen receptors can lead to higher than desired MCAVs. This is supported by the actual values, where MCAV for the bash process is 0.38 in R20 as opposed to 0.22 in R2. This difference is statistically significant, demonstrated through the use of a paired t-test ($p > 0.05$).

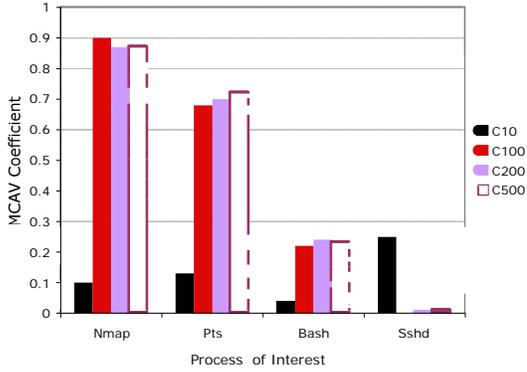

Fig. 10. Cell number MCAV per process for the attack datasets

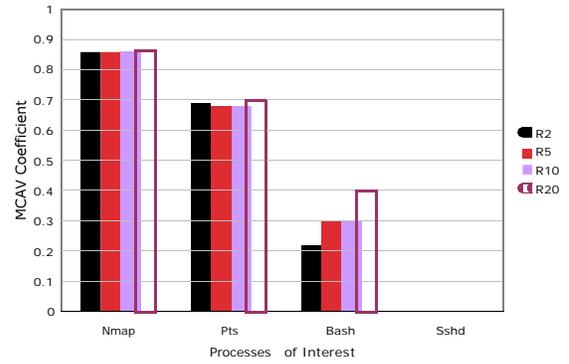

Fig. 12. Number of antigen receptors MCAV per process for the attack datasets

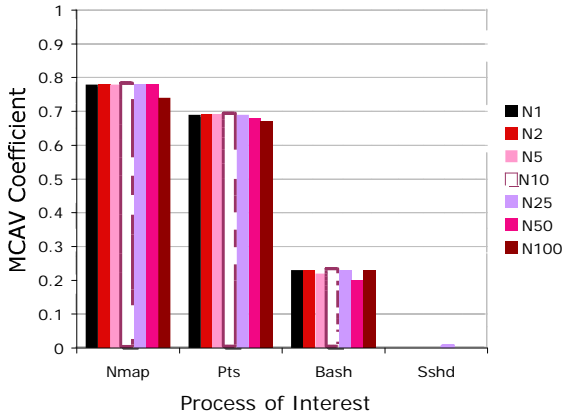

Fig. 11. DC antigen vector size MCAV per process for the attack datasets

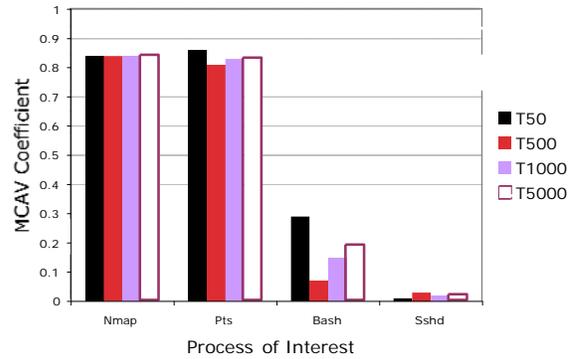

Fig. 13. MCAV per process

The results for the tissue vector size, shown in Figure 13, are similar to the receptor results, in that no significant differences were shown for the nmap, pts and sshd processes. Again, differences were most pronounced for the bash process with a MCAV of 0.28 for T50 and 0.08 for T500. The results show that the DCA is robust to changes in controlling parameters, provided that their values lie within a sensible range.

### 6.9.3. Series-3

Figures 14 to 17 show the MCAVs generated by the weights sensitivity analysis. The resultant surface maps are produced by plotting the two controlling weights, W1 and W2 on the x- and y-axes respectively, and the MCAV per process present on the z-axis. The two anomalous processes are shown in Figure 14 and 15. The surfaces created in these figures show that MCAV values for these processes lie consistently above 0.8. This indicates that the detection of anomalous processes is insensitive to the values of the weights.

Figures 16 and 17 show the results for the sshd and bash processes. Figure 16 exhibits most variation within these four graphs. Peaks of high MCAV in excess of 0.8 are shown when W1 = 1 and W2 < 8. A similar peak is evident in Figure 17. This graph also shows that once both parameters are above 4, the MCAV for the normal processes is small. This implies an effect on the system, which further investigation will clarify.

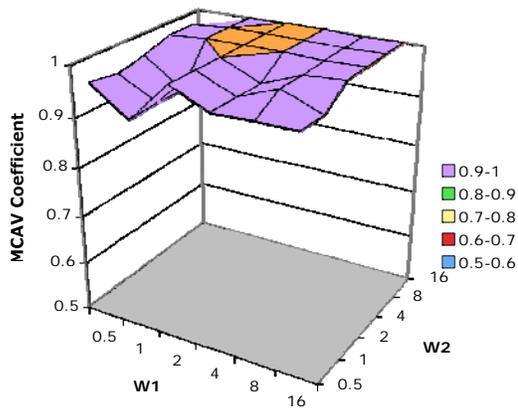

Fig. 14. 3D surface plot of varying weights W1 and W2. Nmap process represented.

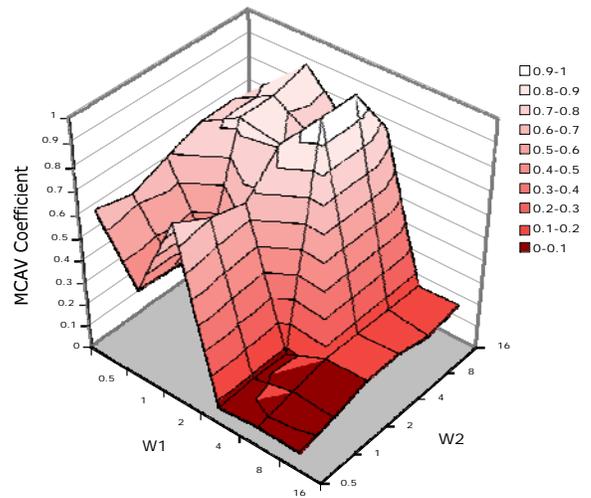

Fig. 16. 3D surface plot of varying weights W1 and W2. Bash process represented.

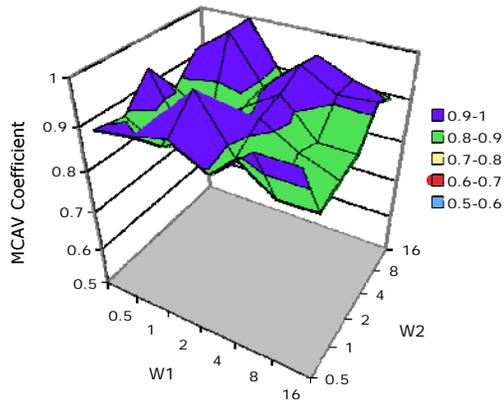

Fig. 15. 3D surface plot of varying weights W1 and W2. Pts process represented.

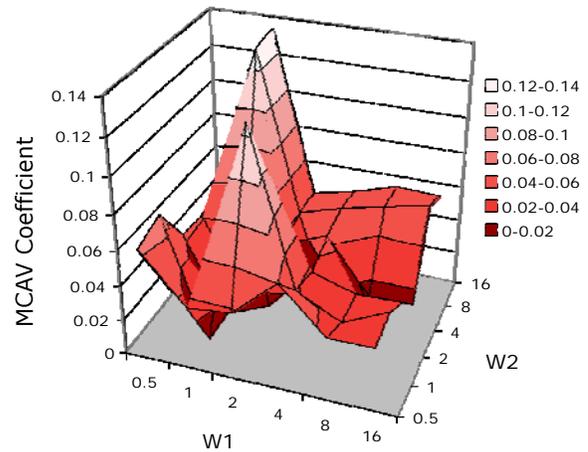

Fig. 17. 3D surface plot of varying weights W1 and W2. Sshd process represented.

## 7. Analysis

In experiment M1 distinct differences are shown in the behaviour of the algorithm for the detection of normal and anomalous processes. The MCAV for the anomalous processes is significantly larger than the MCAV of the normal. This is encouraging as it shows that the DCA can differentiate between two different types of process based on environmentally derived signals. In experiment M2 the PAMP and danger signals were switched. In comparison with the results presented for experiment M1, the MCAV for the anomalous process is not significantly different (paired t-test $p < 0.01$). However, in experiment M2, the standard deviations of the mean MCAVs are generally larger and is especially notable for the nmap process. Potentially, the two signals could be switched (through accidental means or incorrect sig-nal selection) without altering the performance of the algorithm significantly.

Experiment M3 involved reversing the mapping of safe and PAMP signals. The safe signal is generated continuously when the system is inactive and when mapped as a PAMP constantly generated full maturation in the artificial DCs, shown by the high MCAV value for all processes indiscriminately. Interestingly, in M3 the MCAV value for the anomalous processes in the attack datasets is lower than the normal process' value. For the normal dataset, all processes are classified as anomalous, all resulting in a MCAV of 1. Similar impeded performance is shown for M6, caused by the incorrect mapping of a PAMP as a safe signal. The input PAMP signal is strong, yet does not occur throughout. Therefore, not enough suppression is present when the PAMP is mapped.

M5 also produced interesting results - while it did not have such a marked effect on the anomalous processes, it produced high MCAVs for normal items in the attack dataset, but not in the normal dataset. Under 'normal' scenarios this mapping functions as the danger signals are counter-balanced by the safe signals, resulting in low MCAVs. As the PAMP signal is infrequent, insufficient signal to cause full maturation is present.

The intended signal mapping produced good results, showing that the DCA is capable of performing information fusion and anomaly detection. Changing the mapping of signal meaning with data source has shown that the correct mapping is ideal. However, if the PAMP signal is mapped as a danger signal, performance is not sacrificed. Alternatively, PAMPs mapped as safe signals produced the worst results, indicating that care must be taken when selecting a mechanism of suppression. These data also suggest that suppression is a key part of the system, which supports parts of Matzinger's danger theory[19] in reference to peripheral tolerance.

The parameters investigated in series-2 have little influence on the output of the system. For example, varying the DC antigen vector size does not produce any results which are significantly different in this respect. Similar trends are shown for the number of receptors and the number of cells. Exceptions to this include very low values of cells, storage and receptors. The values originally used as default parameters have in many cases produced the most consistent results. This is highlighted in the cell numbers experiments, DC antigen vector size and number of receptors. This is no coincidence as these values, initially derived from biological information[32], and are designed to work together. This may account for some of the robustness seen with these parameters.

Series-3 has provided valuable insight into the behaviour of the DCA. The results in Figures 14 and 15 show that the DCA is insensitive to changes in weights within the signal processing equation, as little variation is shown across the spectrum of values. Significant variation is evident in Figures 16 and 17 suggesting that incorrect weight values may lead to increased values of MCAV for normal processes. The relationship between the two values suggests that higher values for the weight produce lower MCAVs. One reason for this may be linked to the number of update cycles a cell performs. W1 is the controlling weight for output signal $o_0$, which is matched against the DCs migration thresholds. The sooner this threshold is exceeded, the shorter the time a cell spends sampling signals. For this particular dataset, using a threshold of 60 (+/- 50%), a W1 value of over 4 and W2 value of above 8 yields the best result in both cases. This implies that a tighter temporal coupling between signals and antigen produces lower MCAVs for normal processes. To confirm this, a similar analysis will be performed using longer scans in future work.

## 8. Conclusions and Future Work

In this paper the DCA is described in detail and interesting facets of the algorithm are presented. The DCA combines inspiration from the immune system with principles of information fusion to produce an effective anomaly detection technique. The importance of careful signal selection has been highlighted through signal mapping experiment. The DCA is somewhat robust to misrepresentation of the activating danger and PAMP signals, but care must be taken to select a suitable safe signal as an indicator of normality. Incorrect mapping of safe signals can result in impeded performance as shown with our results.

The algorithm has various parameters, and it is shown that the DCA is insensitive to changes in these parameters. Provided that the values are within a sensible range, the algorithm performs well on the task of detecting a ping based port scan. Sensitivity analysis is also performed. The detection of the anomalous processes is robust to changes in the signal processing weights, though large variations are shown in the incorrect detection of normal processes. For the ping scan investigation, larger weights are preferable. This implies that better performance is given if the time spent sampling signals by the DC is shorter.

The DCA is a new development in artificial immune systems, and as yet has not been extensively tested. Its unique methods of combining multiple signals and correlating the combined values with a separate antigen data-stream work well for the purpose of port scan detection. However, this makes the system difficult to compare, as other techniques cannot use data of this type, such as standard machine learning techniques or signature based IDS. Plus, individual signals alone are insufficient to produce classification[10].

The general applicability of the algorithm to a variety of problems is unexplored. This could be initially characterised through the DCA's application to a range of portscans, and then by its applications to other time-dependent datasets. This has thus far included applications within sensor networks, as shown by Kim et al [16]. They used the suggested signal mapping schema as shown in section 5.6.

Future work with the algorithm includes its application to SYN scan detection, where we hope it will produce competitive solutions with other port scan detectors. Such experiment will require the use of multiple signals per category, with a view to a full implementation as a host based port scan detector. The introduction of adaptive signals or variable weights may be necessary once multiple signals per category are used. Although the relative weighs used in the signal processing equation are part of the abstract model, some adaptive mechanism for controlling the values of these weights may be beneficial to the sensitivity of the system. The algorithm may also be applied to other detection or data correlation problems such as the analysis of radio signal data from space, sensor networks, internet worm detection and other security and defence applications.

9. Acknowlegements

This project is supported by the EPSRC, grant number GR/S47809/01. DC photographs provided by Dr Julie McLeod and colleagues at UWE, UK. Graphic design by Mark Hammonds.